\def\year{2020}\relax
\documentclass[letterpaper]{article} 
\usepackage{aaai20}  
\usepackage{times}  
\usepackage{helvet} 
\usepackage{courier}  
\usepackage{comment}
\usepackage[hyphens]{url}  
\usepackage{graphicx} 
\usepackage{subcaption}
\usepackage{graphicx}  

\title{When AWGN-based Denoiser Meets Real Noises}
\author{
{Yuqian Zhou$^1$, Jianbo Jiao$^{2}$\thanks{Corresponding author}, Haibin Huang$^3$, Yang Wang$^4$, Jue Wang$^{^3}$, Honghui Shi$^{1}$, Thomas Huang$^{1}$}\\
{\small $^{1}$IFP Group, UIUC~~~ $^{2}$University of Oxford ~~~ $^{3}$Megvii Research ~~~ $^{4}$Stony Brook}\\
{\tt\small \{yuqian2, t-huang1\}@illinois.edu}
}

\begin{document}

\maketitle

\begin{abstract}
Discriminative learning based image denoisers have achieved promising performance on synthetic noises such as Additive White Gaussian Noise (AWGN). The synthetic noises adopted in most previous work are pixel-independent, but real noises are mostly spatially/channel-correlated and spatially/channel-variant. This domain gap yields unsatisfied performance on images with real noises if the model is only trained with AWGN. In this paper, we propose a novel approach to boost the performance of a real image denoiser which is trained only with synthetic pixel-independent noise data dominated by AWGN. First, we train a deep model that consists of a noise estimator and a denoiser with mixed AWGN and Random Value Impulse Noise (RVIN). We then investigate Pixel-shuffle Down-sampling (PD) strategy to adapt the trained model to real noises. Extensive experiments demonstrate the effectiveness and generalization of the proposed approach. Notably, our method achieves state-of-the-art performance on real sRGB images in the DND benchmark among models trained with synthetic noises. Codes are available at \url{https://github.com/yzhouas/PD-Denoising-pytorch}.
\end{abstract}

\section{Introduction}
\begin{figure}[t]
	\begin{center}
		\includegraphics[width=1\linewidth]{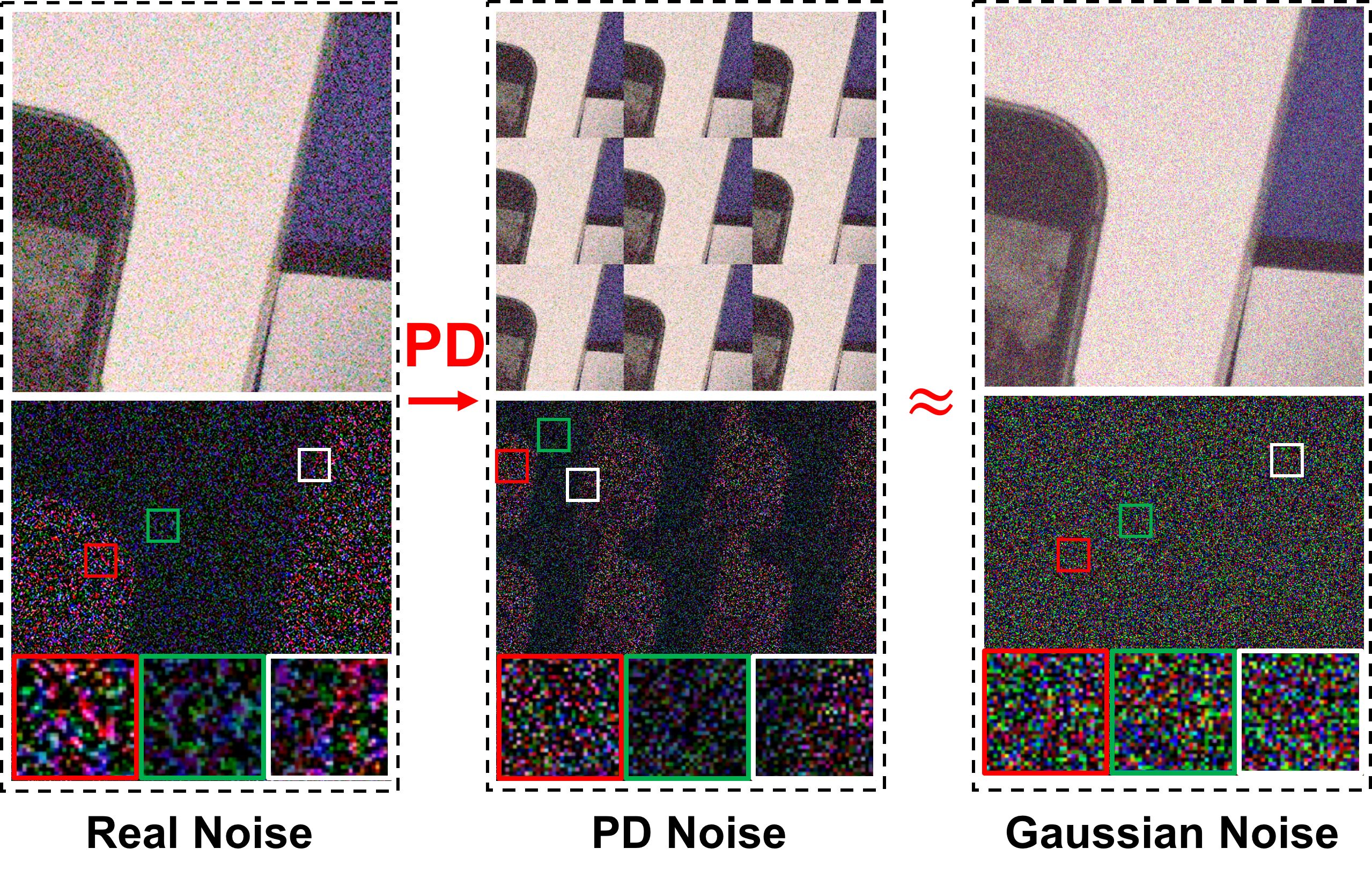}
	\end{center}
	\caption{Basic idea of the proposed adaptation method: Pixel-shuffle Down-sampling (PD). Spatially-correlated real noise (Left) is broken into spatially-variant pixel-independent noise (Middle) to approximate spatially-variant Gaussian noise (Right). Then an AWGN-based denoiser can be applied to such real noise accordingly.}
	\label{fig:noise_comp}
\end{figure}
As a fundamental task in image processing and computer vision,  image denoising has been extensively explored in the past several decades even for downstream applications~\cite{zhou2018survey,wang2019video}. Traditional methods including the ones based on image filtering ~~\cite{dabov2008image}, low rank approximation ~\cite{gu2014weighted,xu2017multi,yair2018multi}, sparse coding ~\cite{elad2006image}, and image prior~\cite{ulyanov2017deep} have achieved satisfactory results on synthetic noise such as Additive White Gaussian Noise (AWGN). Recently, deep CNN has been applied to this task, and discriminative-learning-based methods such as DnCNN~\cite{zhang2017beyond} outperform most traditional methods on AWGN denoising.

Unfortunately, while these learning-based methods work well on the same type of synthetic noise that they were trained on, their performance degrades rapidly on real images, showing poor generalization ability in real world applications. This indicates that these data-driven denoising models are highly domain-specific and non-flexible to transfer to other noise types beyond AWGN. To improve model flexibility, the recently-proposed FFDNet~\cite{zhang2018ffdnet} trains a conditional non-blind denoiser with a manually adjusted noise-level map. By giving high-valued uniform maps to FFDNet, only over-smoothed results can be obtained in real image denoising. Therefore, blind denoising of real images is still very challenging due to the lack of accurate modeling of real noise distribution. These unknown real-world noises are much more complex than pixel-independent AWGN. They can be  spatially-variant, spatially-correlated, signal-dependent, and even device-dependent. 

To better address the problem of real image denoising, current attempts can be roughly divided into the following categories: (1) realistic noise modeling ~\cite{guo2018toward,brooks2019unprocessing,abdelhamed2019ntire}, (2) noise profiling such as multi-scale ~\cite{lebrun2015multiscale,yair2018multi}, multi-channel ~\cite{xu2017multi} and regional based ~\cite{liu2017image} settings, and (3) data augmentation techniques such as the adversarial-learning-based ones ~\cite{chen2018image}. Among them, CBDNet ~\cite{guo2018toward} achieves good performance by modeling the realistic noise using the in-camera pipeline model proposed in ~\cite{liu2008automatic}. It also trains an explicit noise estimator and sets a larger penalty for under-estimated noise. The network is trained on both synthetic and real noises, but it still cannot fully characterize real noises. Brooks et al. \cite{brooks2019unprocessing} used prior statistics stored in the raw data of DND to augment the synthetic RGB data, but it does not prove the generalization of the model on other real noises. 

In this work, from a novel viewpoint of real image blind denoising, we seek to adapt a learning-based denoiser trained on pixel-independent synthetic noises to unknown real noises. As shown in Figure \ref{fig:noise_comp}, we assume that real noises differ from pixel-independent synthetic noises dominantly in {\em spatial/channel-variance and correlation}~\cite{DD}. This difference results from in-camera pipeline like demosaicing~\cite{zhou2019adaptation}. Based on this assumption, we first propose to train a basis denoising network using mixed AWGN and RVIN. Our flexible basis net consists of an explicit noise estimator followed by a conditional denoiser. We demonstrate that this fully-convolutional nets are actually efficient in coping with pixel-independent spatially/channel-variant noises. Second, we propose a simple yet effective adaptation strategy, Pixel-shuffle Down-sampling(PD), which employs the divide-and-conquer idea to handle real noises by breaking down the spatial correlation.  

In summary, our main contributions include:
\begin{itemize}

\item
We propose a new flexible deep denoising model (trained with AWGN and RVIN) for both blind and non-blind image denoising. We also demonstrate that such fully convolutional models trained on spatially-invariant noises can handle {\em spatially-variant noises}. 
\item
We adapt the AWGN-RVIN-trained deep denoiser to real noises by applying a novel strategy called Pixel-shuffle Down-sampling (PD). {\em Spatially-correlated noises} are broken down to {\em pixel-wise independent noises}. We examine and overcome the proposed domain gap to boost real denoising performance.
\item
The proposed method achieves state-of-the-art performance on DND benchmark and other real noisy RGB images among models trained only with synthetic noises. Note that our model does not use any images or prior meta-data from real noise datasets. We also show that with the proposed PD strategy, the performance of some other existing denoising models can also be boosted. 
\end{itemize}

\section{Related Work}
\textbf{Discriminative Learning based Denoiser.}
Denoising methods based on CNNs have achieved impressive performance on removing synthetic Gaussian noise.  Burger et al. ~\cite{burger2012image} proposed to apply multi-layer perceptron (MLP) to denoising task. In ~\cite{chen2017trainable}, Chen et al. proposed  a trainable nonlinear reaction diffusion (TNRD) model for Gaussian noise removal at different level. 
DnCNN ~\cite{zhang2017beyond} was the first to propose a blind Gaussian denoising network using deep CNNs. It demonstrated the effectiveness of residual learning and batch normalization. More network structures like dilated convolution ~\cite{zhang2017learning}, autoencoder with skip connection ~\cite{mao2016image}, ResNet ~\cite{ren2018dn}, recursively branched deconvolutional network (RBDN) ~\cite{santhanam2017generalized} were proposed to either enlarge the receptive field or balance the efficiency. Recently some interests are put into combining image denoising with high-level vision tasks like classification and segmentation. Liu et al. ~\cite{liu2017image} applied segmentation to enhance the denoising performance on different regions. Similar class-aware work were developed in ~\cite{niknejad2017class}. Due to domain-specific training and deficient realistic noise data, those deep models are not robust enough on realistic noises. In recently proposed FFDNet ~\cite{zhang2018ffdnet}, the author proposed a non-blind denoising by concatenating the noise level as a map to the noisy image. By manually adjusting noise level to a higher value, FFDNet demonstrates a spatial-invariant denoising on realistic noises with over-smoothed details.

\textbf{Blind Denoising on Real Noisy Images.}
Real noises of CCD cameras are complicated and are related to optical sensors and in-camera process. Specifically, multiple noise sources like photon noise, read-out noise etc. and processing including demosaicing, color and gamma transformation introduce the main characteristics of real noises: spatial/channel correlation, variance, and signal-dependence. To approximate real noise, multiple types of synthetic noise are explored in previous work, including Gaussian-Poisson ~\cite{foi2008practical,liu2014practical}, Gaussian Mixture Model (GMM) ~\cite{zhu2016noise}, in-camera process simulation ~\cite{liu2008automatic,guo2018toward} and GAN-generated noises ~\cite{chen2018image}, to name a few. CBDNet ~\cite{guo2018toward} first simulated real noise and trained a subnetwork for noise estimation, in which spatial-variance noise is represented as spatial maps. Besides, multi-channel~\cite{xu2017multi,guo2018toward} and multi-scale~\cite{lebrun2015multiscale,yu2015multi} strategy were also investigated for adaptation. Different from all the aforementioned works which focus on directly synthesizing or simulating noises for training, in this work, we apply AWGN-RVIN model and focus on pixel-shuffle adaptation strategy to fill in the gap between pixel-independent synthetic and pixel-correlated real noises. 
\section{Methodology}
\subsection{Basis Noise Model}
\begin{figure}[t]
	\centering
		\includegraphics[width=\linewidth]{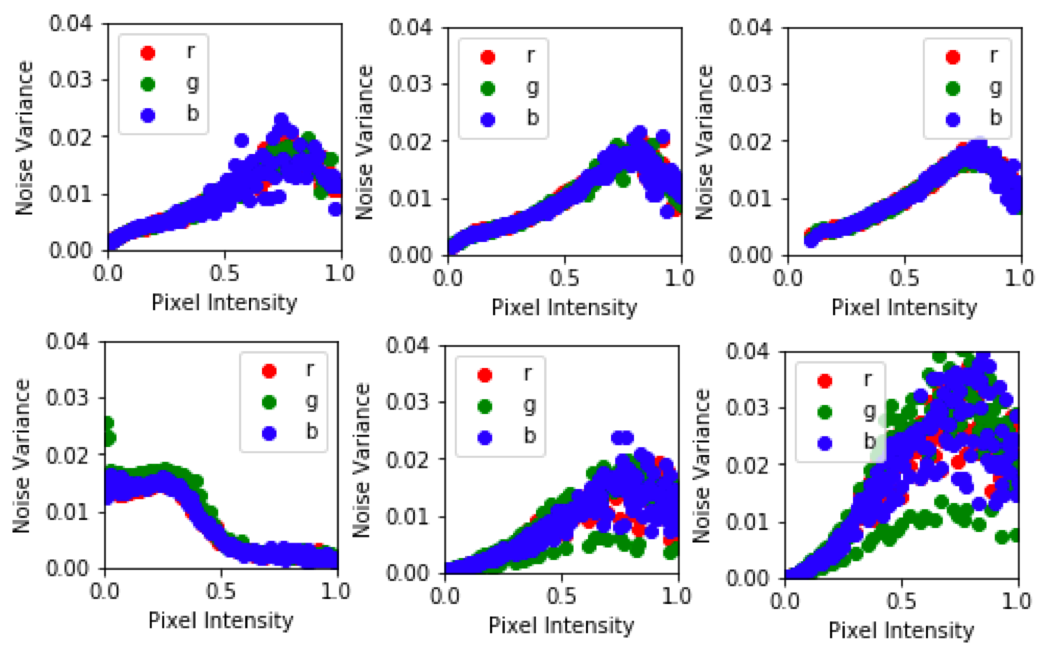}
	\caption{Noise Level Function (NLFs) (noise variance as a function of image intensity) before (first row) and after (second row) gamma transform and demosaicing. Gamma factor is 0.39, 1.38 and 2.31 from the left to right column. 
	}
	\label{fig:nd}
\end{figure}
The basis noise model is mixed AWGN-RVIN. Noises in sRGB images are no longer approximated Gaussian-Poisson Noises as in the raw sensor data mainly due to gamma transform, demosaicing, and other interpolations etc.. In Figure~\ref{fig:nd}, we follow~\cite{liu2008automatic} pipeline to synthesize noisy images, and plot the Noise Level Functions (NLFs) (noise variance as a function of image intensity) before (first row) and after (second row) the Gamma Correction transform and demosaicing. From left to right, the Gamma factor increases. It shows that in RGB images, clipping effect and other non-linear transforms will greatly influence the originally linear noise variance-intensity relationship in raw sensor data, even change the noise mean. Tough complicated, for a more general case than Gaussian-Poisson noises of modeling different nonlinear transforms, real noises in RGB can still be locally approximated as AWGN \cite{zhang2018ffdnet,lee1980refined,xu2018trilateral}. In this paper, we thus assume the RGB noises to be approximated spatially-variant and spatially-correlated AWGN.

Adding RVIN for training aims at explicitly resolving the defective pixels caused by dead pixels of camera hardware or long exposure frequently appearing in most night-shot images. We generate AWGN, RVIN and mixed AWGN-RVIN following PGB\cite{xu2016patch}. 

\subsection{Basis Model Structure}
\begin{figure}[t]
	\begin{center}
		\includegraphics[width=1\linewidth]{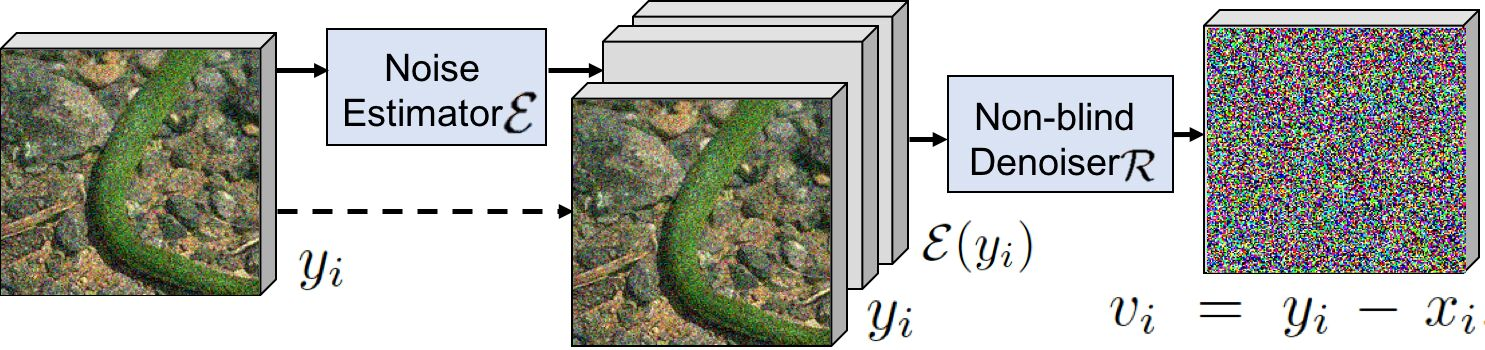}
	\end{center}
	\caption{Structure of the proposed blind denoising model. It consists of a noise estimator $\mathcal{E}$ and a follow-up non-blind denoiser $\mathcal{R}$. The model aims to jointly learn the image residual. }
	\label{fig:structure}
\end{figure}

The architecture of the proposed basis model is illustrated in Figure \ref{fig:structure}. 
The proposed blind denoising model $\mathcal{G}$ consists of a noise estimator $\mathcal{E}$ and a follow-up non-blind denoiser $\mathcal{R}$. Given a noisy observation $y_i = \mathcal{F}(x_i)$, where $\mathcal{F}$ is the noise synthetic process, and $x_i$ is the noise-free image, the model aims to jointly learn the residual $\mathcal{G}(y_i)\approx v_i=y_i-x_i$, and it is trained on paired synthetic data $(y_i, v_i)$. Specifically, the noise estimator outputs $\mathcal{E}(y_i)$ consisting of six pixel-wise noise-level maps that correspond to two noise types, i.e., AWGN and RVIN, across three channels (R, G, B). Then $y_i$ is concatenated with the estimated noise level maps $\mathcal{E}(y_i)$ and fed into the non-blind denoiser $\mathcal{R}$. The denoiser then outputs the noise residual $\mathcal{G}(y_i)=\mathcal{R}(y_i,\mathcal{E}(y_i))$. Three objectives are proposed to supervise the network training, including the noise estimation ($\mathcal{L}_{e}$), blind ($\mathcal{L}_{b}$) and non-blind ($\mathcal{L}_{nb}$) image denoising objectives, defined as, 

\begin{equation}
	\mathcal{L}_{e} = \frac{1}{2N} \sum_{i=1}^{N}{{||\mathcal{E}(y_i; \Theta_{E})-e_i||}_{F}^2},
\label{eq:loss1}
 \end{equation}
 	\vspace{-4mm}
 \begin{equation}
	\mathcal{L}_{b} = \frac{1}{2N} \sum_{i=1}^{N}{{||\mathcal{R}(y_i,\mathcal{E}(y_i;\Theta_{E});\Theta_{R})-v_i||}_{F}^2},
\label{eq:loss2}
 \end{equation}
	\vspace{-4mm}
\begin{equation}
	\mathcal{L}_{nb} = \frac{1}{2N} \sum_{i=1}^{N}{{||\mathcal{R}(y_i,e_i;\Theta_{R})-v_i||}_{F}^2},
\label{eq:loss3}
 \end{equation}
where $\Theta_E$ and $\Theta_R$ are the trainable parameters of $\mathcal{E}$ and $\mathcal{R}$. $e_i$ is the ground truth noise level maps for $y_i$, consisting of ${e_i}_{AWGN}$ and ${e_i}_{RVIN}$. For AWGN, ${e_i}_{AWGN}$ is represented as the even maps filled with the same standard deviation values ranging from 0 to 75 across R,G,B channels. For RVIN, ${e_i}_{RVIN}$ is represented as the maps valued with the corrupted pixels ratio with upper-bound set to 0.3. We further normalize $e_i$ to range [0,1]. Then the full objective can be represented as a weighted sum of the above three losses, 
\begin{equation}
	\mathcal{L} = \alpha \mathcal{L}_{e}  + \beta \mathcal{L}_{b} + \gamma \mathcal{L}_{nb},
\label{eq:full_obj} 
\end{equation}
in which $\alpha$, $\beta$ and $\gamma$ are hyper-parameters to balance the losses, and we set them to be equal for simplicity. 

The proposed model structure can perform both blind and non-blind denoising simultaneously, and the model is more flexible in interactive denoising and result adjustment. Explicit noise estimation also benefits noise modeling and disentanglement.  

\subsection{Pixel-shuffle Down-sampling (PD) Adaptation} 
\paragraph{Pixel-shuffle Down-sampling.} 
Pixel-shuffle \cite{shi2016real} down-sampling is defined to create the mosaic by sampling the images with stride $s$. Compared to other down-sampling methods like linear interpolation, bi-cubic interpolation, and pixel area relation, the pixel-shuffle and nearest-neighbour down-sampling on noisy image would not influence the real noise distribution. Besides, pixel-shuffle also benefits image recovery by preserving the original pixels from the images compared to others. These two advantages yield the two stages of PD strategy: adaptation and refinement.
\paragraph{Adaptation.} 

\begin{figure}[t]
	\centering
	\begin{subfigure}[t]{\linewidth} 
		\includegraphics[width=\textwidth]{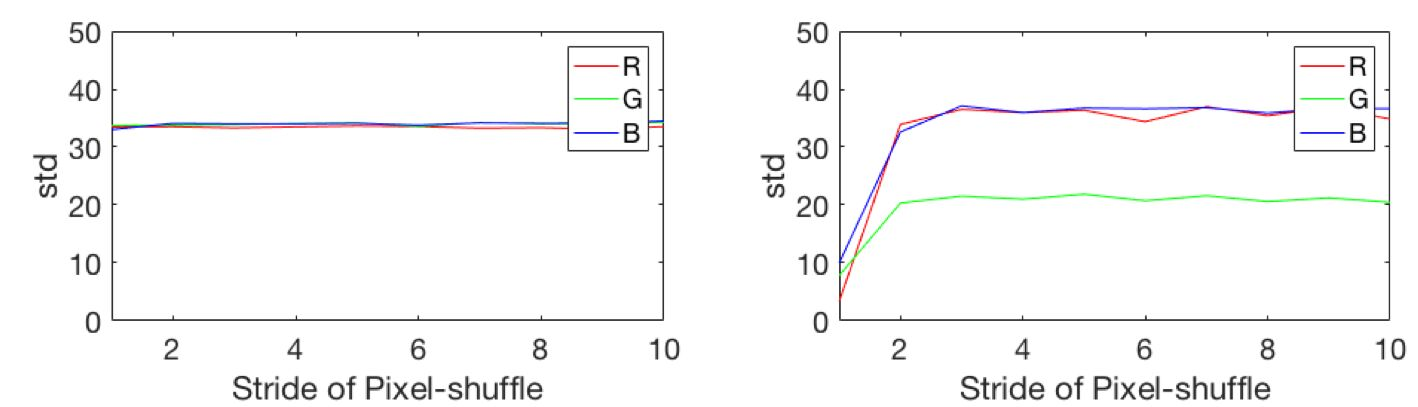}
		\caption{ As the stride increases, Left: Estimated noise level on AWGN-corrupted image. Right: Estimated noise level on real noisy images. } 
	\end{subfigure}
	\begin{subfigure}[t]{\linewidth} 
		\includegraphics[width=\textwidth]{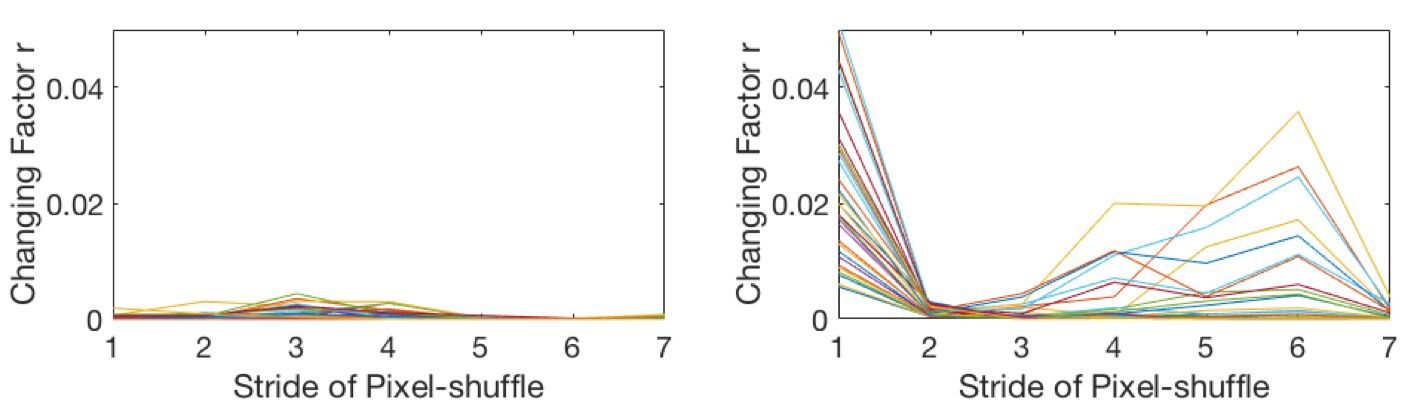}
		\caption{Left: Changing factor $r_s$ on AWGN-corrupted images of CBSD68 and Right: on real noisy images of DND. Different color lines represent different image samples.} 
	\end{subfigure}
	\caption{Influence of Pixel-shuffle on noise patterns and noise estimation algorithms.}
	\label{fig:ps_est}
\end{figure}

Learning-based denoiser trained on AWGN is not robust enough to real noises due to domain difference. To adapt the noise model to real noise, here we briefly analyze and justify our assumption on the difference between real noises and Gaussian noise: spatial/channel variance and correlation. 

Suppose a noise estimator is robust, which means it can accurately estimate the exact noise level, for a single AWGN-corrupted image, pixel-shuffle down-sampling will neither influence the AWGN variance nor the estimation values, when the sample stride is small enough to preserve the textural structures. When extending it to real noise case, we have an interesting hypothesis: as we increase the sample stride of pixel-shuffle, the estimation values of specific noise estimators will first fluctuate and then keep steady for a couple of stride increment. This assumption is feasible because pixel-shuffle will break down the spatial-correlated noise patterns to pixel-independent ones, which can be approximated as spatial-variant AWGN and adapted to those estimators. 

We justify this hypothesis on both \cite{liu2013single} and our proposed pixel-wise estimator. As shown in Figure~\ref{fig:noise_comp}, we randomly cropped a patch of size $200 \times 200$ from a random noisy image $y$ in SIDD\cite{abdelhamed2018high}. We add AWGN with $std=35$ to its noise-free ground truth $x$. After pixel-shuffling both $y$ and AWGN-corrupted $x$, starting from stride $s=2$, the noise pattern of $y$ demonstrates expected pixel independence. Using \cite{liu2013single}, the estimation result for $x$ is unchanged in Figure \ref{fig:ps_est} (a) (Left), but the one for $y$ in Figure \ref{fig:ps_est} (a) (Right) first increases and begins to keep steady after stride $s=2$. It is consistent with the visual pattern and our hypothesis.

One assumption of \cite{liu2013single} is that the noise is additive and evenly distributed across the image. For spatial-variant signal-dependent real noises, our pixel-wise estimator has its superiority. To make statistics of spatial-variant noise estimation values, we extract the three AWGN channels of noise map $\mathcal{E}_{AWGN}(y_i) \in R^{W \times H \times 3}$, where $W$ and $H$ are width and height of the input image, and compute the normalized 10-bin histograms $h_s \in R^{10 \times 3}$ across each channel when the stride is $s$. We introduce the changing factor $r_{s}$ to monitor the noise map distribution changes as the stride $s$ increases,   
\begin{equation}
	r_{s} = E_c||h_{sc}-h_{(s+1)c}||^2_2,
\label{eq:r} 
 \end{equation}
where $c$ is the channel index. We then investigate the difference of $r_{s}$ sequence between AWGN and realistic noises. Specifically, we randomly select 50 images from CBSD68 \cite{roth2009fields} and add random-level AWGN to them. For comparison, we randomly pick up 50 image patches of $512\times512$ from DND benchmark. In Figure \ref{fig:ps_est} (b), $r_{s}$ sequence remains closed to zero for all AWGN-currupted images (Left figure), while for real noises $r_{\alpha}$ demonstrates an abrupt drop when $s=2$. It indicates that the spatial-correlation has been broken from $s=2$. 

The above analysis inspires the proposed adaptation strategy based on pixel-shuffle. Intuitively, we aim at finding the smallest stride $s$ to make the down-sampled spatial-correlated noises match the pixel-independent AWGN. Thus we keep increasing the stride $s$ until $r_{s}$ drops under a threshold $\tau$. We run the above experiments on CBSD68 for 100 iterations to select the proper generalized threshold $\tau$. After averaging the maximum $r$ of each iteration, we empirically set $\tau = 0.008$.

\begin{figure}[t]
	\begin{center}
		\includegraphics[width=1\linewidth]{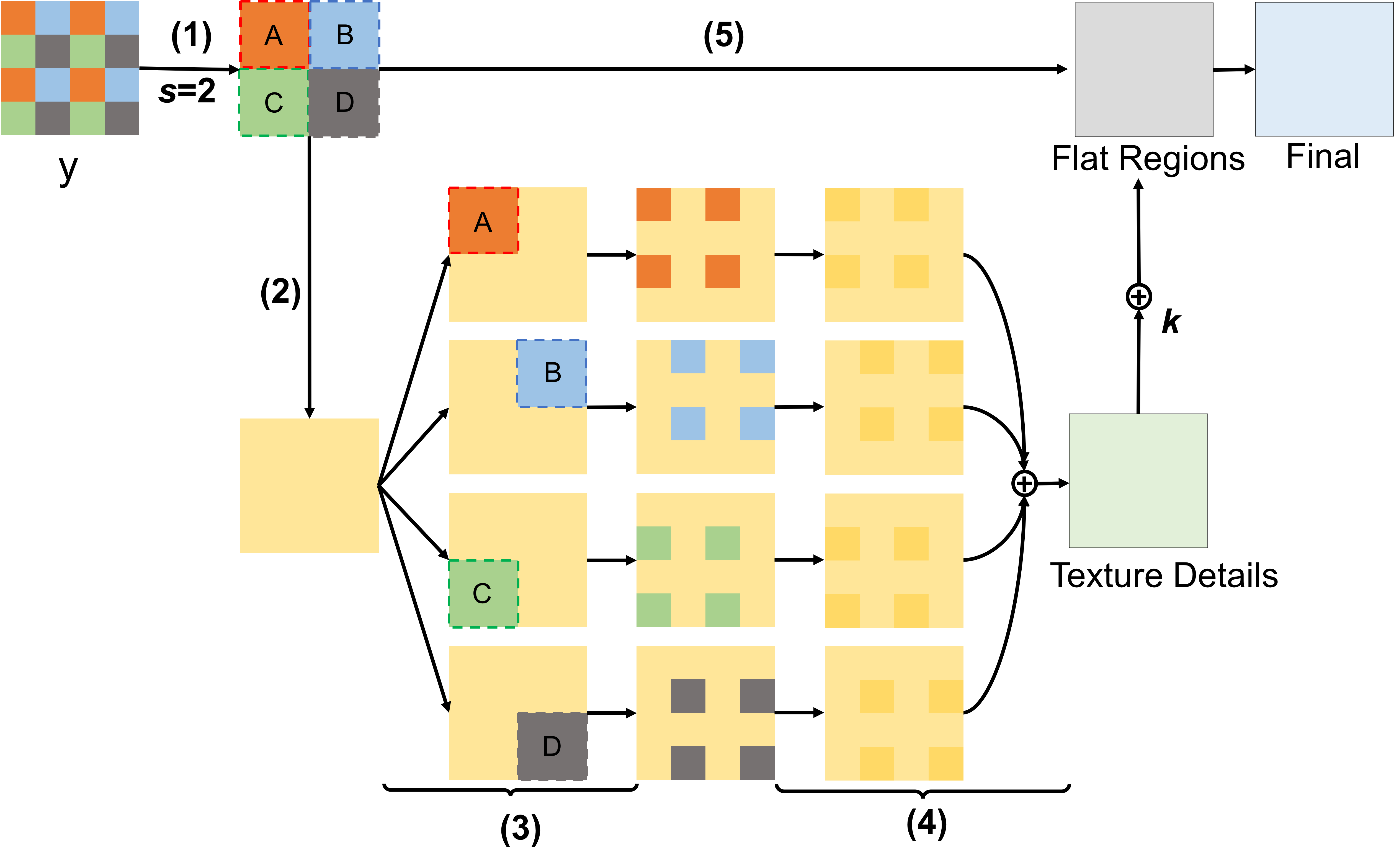}
	\end{center}
	\vspace{-6mm}
	\caption{Pixel-shuffle Down-sampling (PD) refinement strategy with $s=2$.}
	\label{fig:ps_pipeline}
\end{figure}

\paragraph{PD Refinement.} 
Figure \ref{fig:ps_pipeline} shows the proposed Pixel-shuffle Down-sampling (PD) refinement strategy: (1) Compute the smallest stride $s$, which is 2 in this example and more digital camera image cases, to match 
AWGN following the adaptation process, and pixel-shuffle the image into mosaic $y_{s}$; (2) Denoise $y_{s}$ using $\mathcal{G}$; (3) Refill each sub-image with noisy blocks separately and pixel-shuffle upsample them; (4) Denoise each refilled image again using $\mathcal{G}$ and average them to obtain the `texture details' $T$; (5) Combine the over-smoothed `flat regions' $F$ to refine the final result. 

As summarized in \cite{liu2008automatic}, the goals of noise removal include preserving texture details and boundaries, smoothing flat regions, and avoiding generating artifacts. Therefore, in the above step-(5), we propose to further refine the denoised image with the combination of `texture details' $T$ and `flat regions' $F$. `Flat regions' can be obtained from over-smoothed denoising results generated by lifting the noise estimation levels. In this work, given a noisy observation $y$, the refined noise maps are defined as,
\begin{equation}
\resizebox{.9\hsize}{!}{$
  \hat{\mathcal{E}(PD(y))}(i,j) = \max_{i,j}{\mathcal{E}(PD(y))(i,j)}, i \in [1,W], j\in [1,H].
  $}
  \label{equ:max}
\end{equation}

Consequently, the `flat region' is defined as  $F =PU( \mathcal{R}(PD(y), \hat{\mathcal{E}(PD(y))}))$, where PD and PU are pixel-shuffle downsampling and upsampling. The final result is obtained by $kF + (1-k)T$.

\section{Experiments}
\subsection{Implementation Details}
In this work, the structures of the sub-network $\mathcal{E}$ and $\mathcal{R}$ follow DnCNN~\cite{zhang2017beyond} of 5 layers and 20 layers. For grayscale image experiments, we also follow DnCNN to crop $50 \times 50$ patches from 400 images of size $180 \times 180$. For color image model, we crop $50 \times 50$ patches with stride 10 from 432 color images in the Berkeley segmentation dataset (BSD)~\cite{roth2009fields}. The training data ratio of single-type noises (either AWGN or RVIN) and mixed noises (AWGN and RVIN) is 1:1. During training, Adam optimizer is utilized and the learning rate is set to $10^{-3}$, and batch size is 128. After 30 epochs, the learning rate drops to $10^{-4}$ and the training stops at epoch 50.

To evaluate the algorithm on synthetic noise (AWGN, mixed AWGN-RVIN and spatially-variant Gaussian), we utilize the benchmark data from BSD68, Set20~\cite{xu2016patch} and CBSD68 \cite{roth2009fields}. For realistic noise, we test it on RNI15~\cite{RNI15}, DND benchmark~\cite{plotz2017benchmarking}, and self-captured night photos. We evaluate the performance of the algorithm in terms of PSNR and SSIM. Qualitative performance for denoising is also presented, with comparison to other state-of-the-arts. 
\subsection{Evaluation with Synthetic Noise}
\begin{figure*}[t]
	\centering
	\begin{subfigure}{0.24\linewidth} 
		\includegraphics[width=\textwidth]{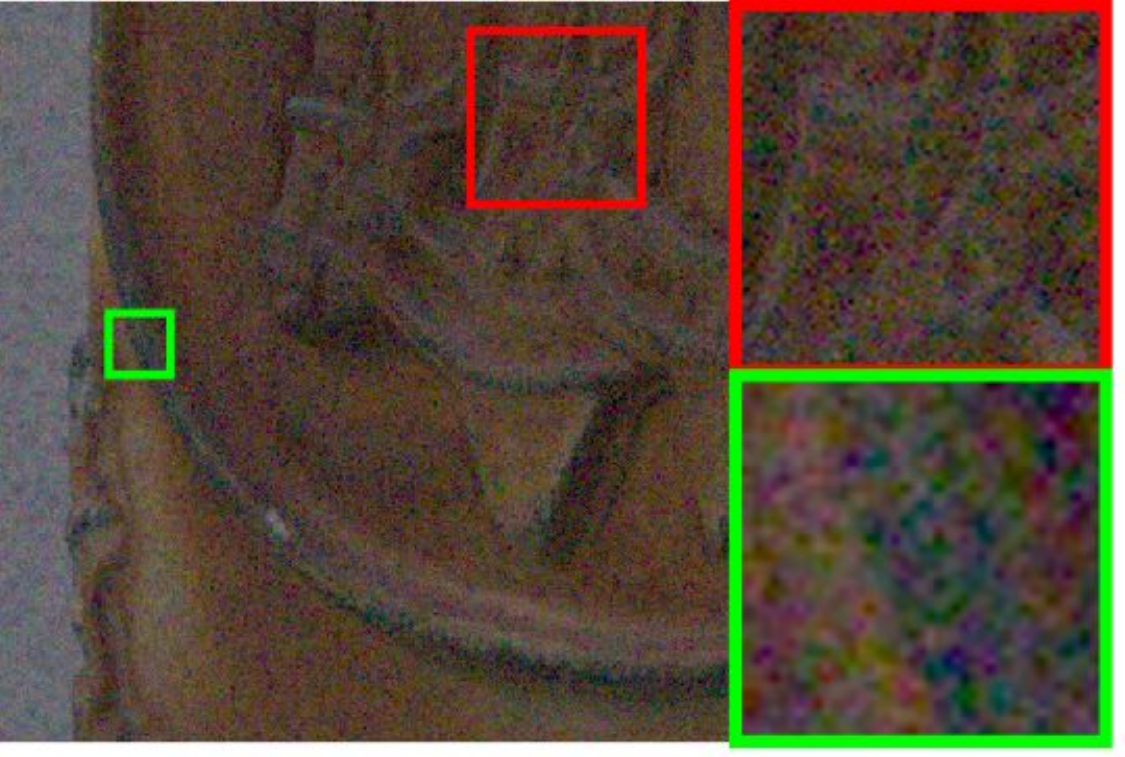}
		\caption{Noisy Image} 
	\end{subfigure}
	\begin{subfigure}{0.24\linewidth} 
		\includegraphics[width=\textwidth]{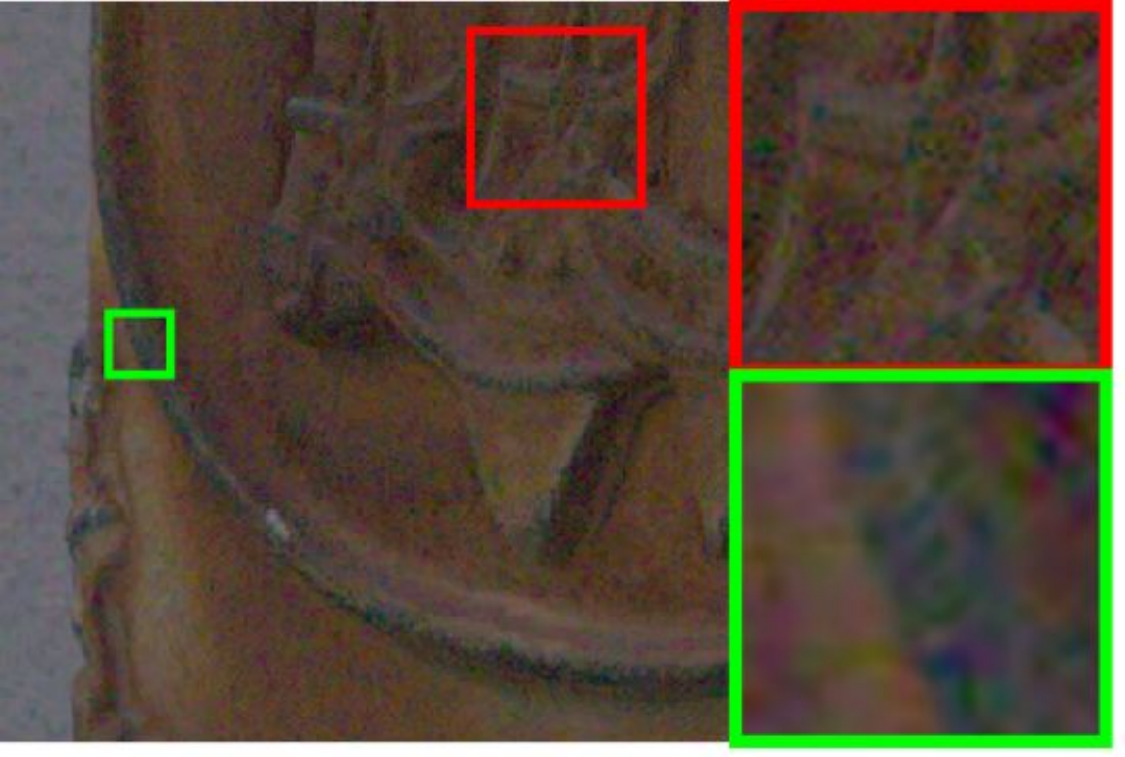}
		\caption{CBM3D(29.33dB)} 
	\end{subfigure}
	\begin{subfigure}{0.24\linewidth} 
		\includegraphics[width=\textwidth]{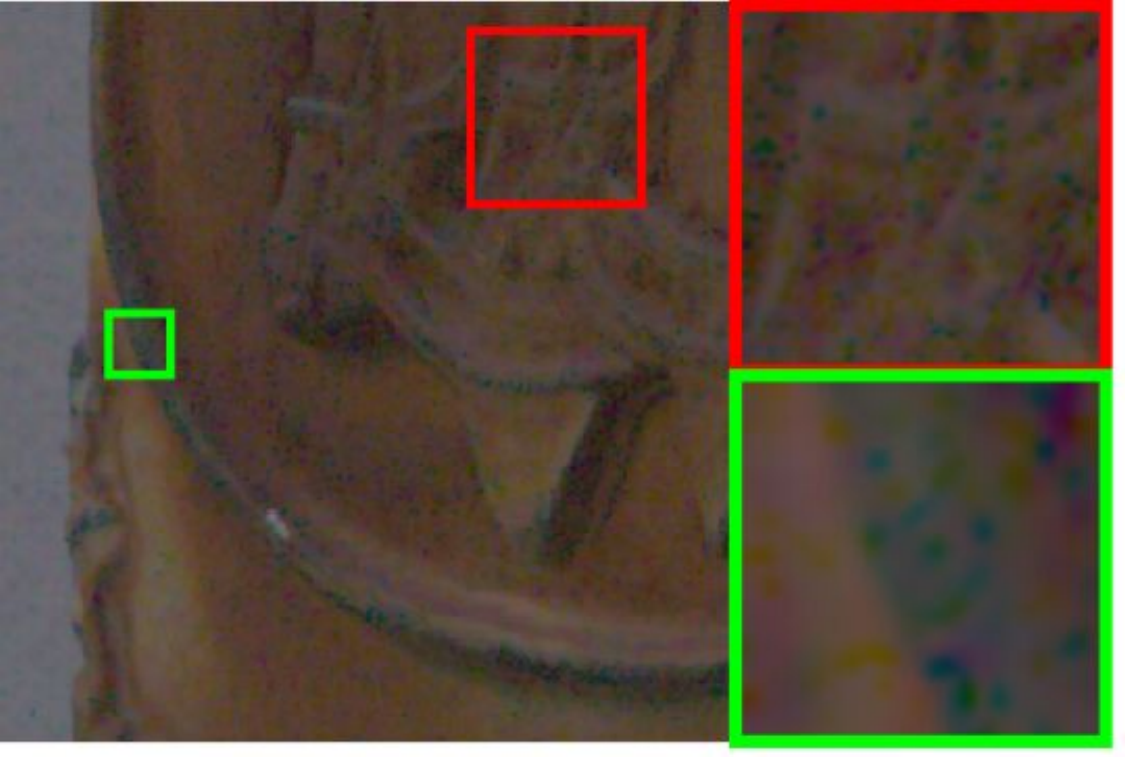}
		\caption{WNNM(29.80dB)} 
	\end{subfigure}
	\begin{subfigure}{0.24\linewidth} 
		\includegraphics[width=\textwidth]{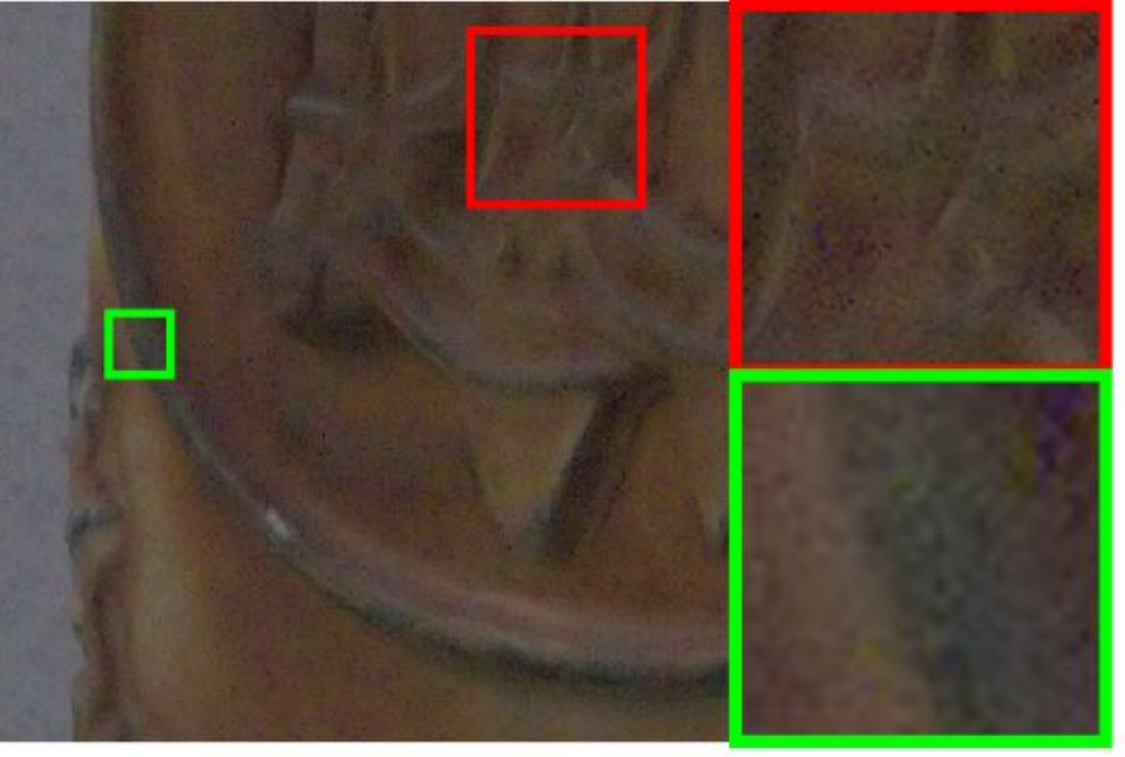}
		\caption{NI(32.29dB)} 
	\end{subfigure}
	\begin{subfigure}{0.24\linewidth} 
		\includegraphics[width=\textwidth]{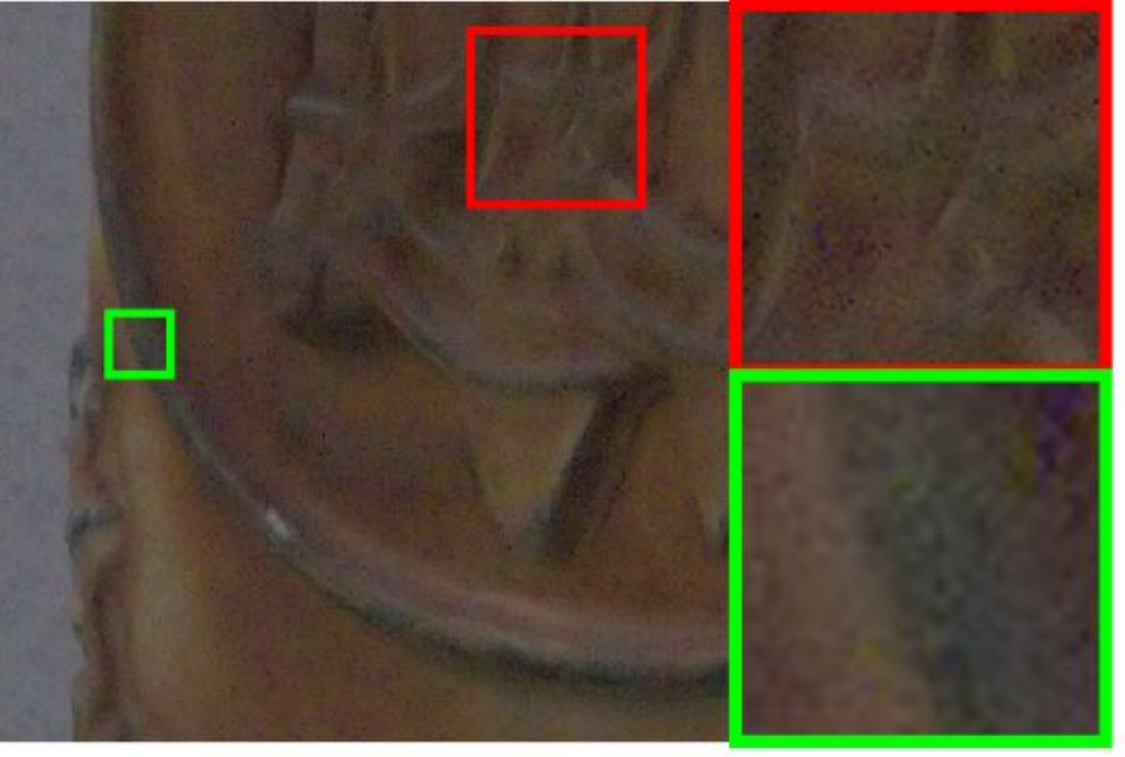}
		\caption{NC(32.29dB)} 
	\end{subfigure}
	\begin{subfigure}{0.24\linewidth} 
		\includegraphics[width=\textwidth]{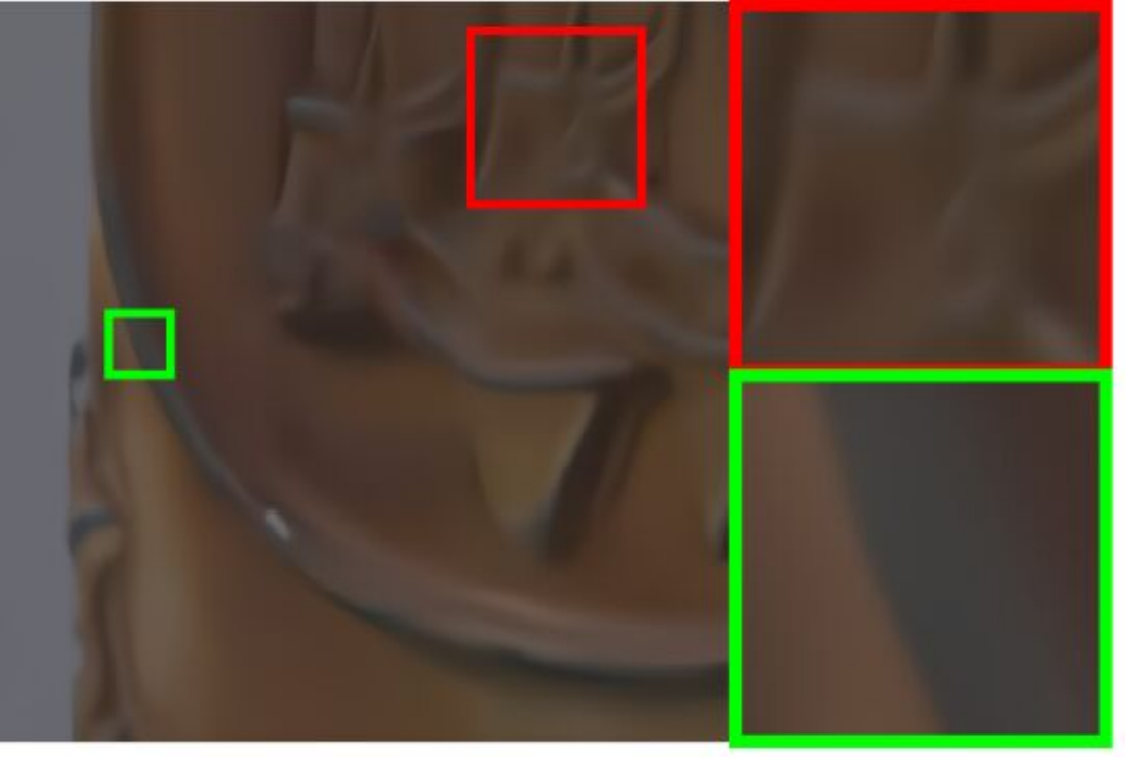}
		\caption{FFDNet(34.47dB)} 
	\end{subfigure}
	\begin{subfigure}{0.24\linewidth} 
		\includegraphics[width=\textwidth]{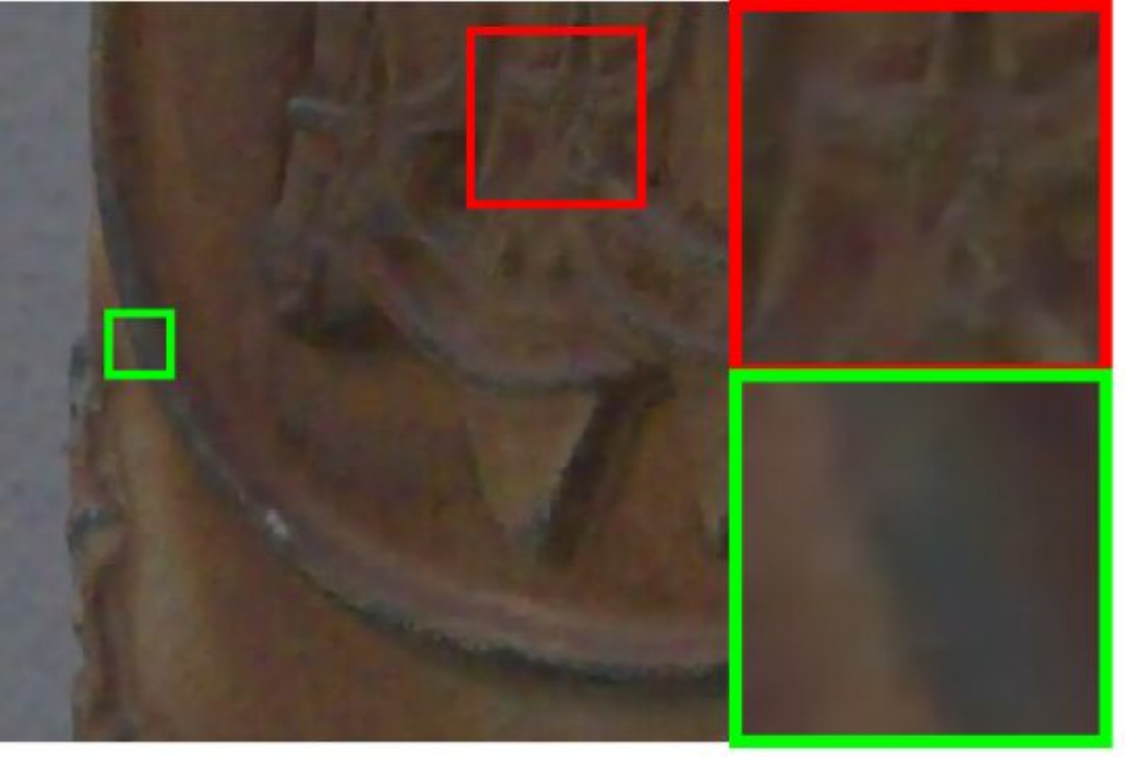}
		\caption{CBDNet(34.50dB)} 
	\end{subfigure}
	\begin{subfigure}{0.24\linewidth} 
		\includegraphics[width=\textwidth]{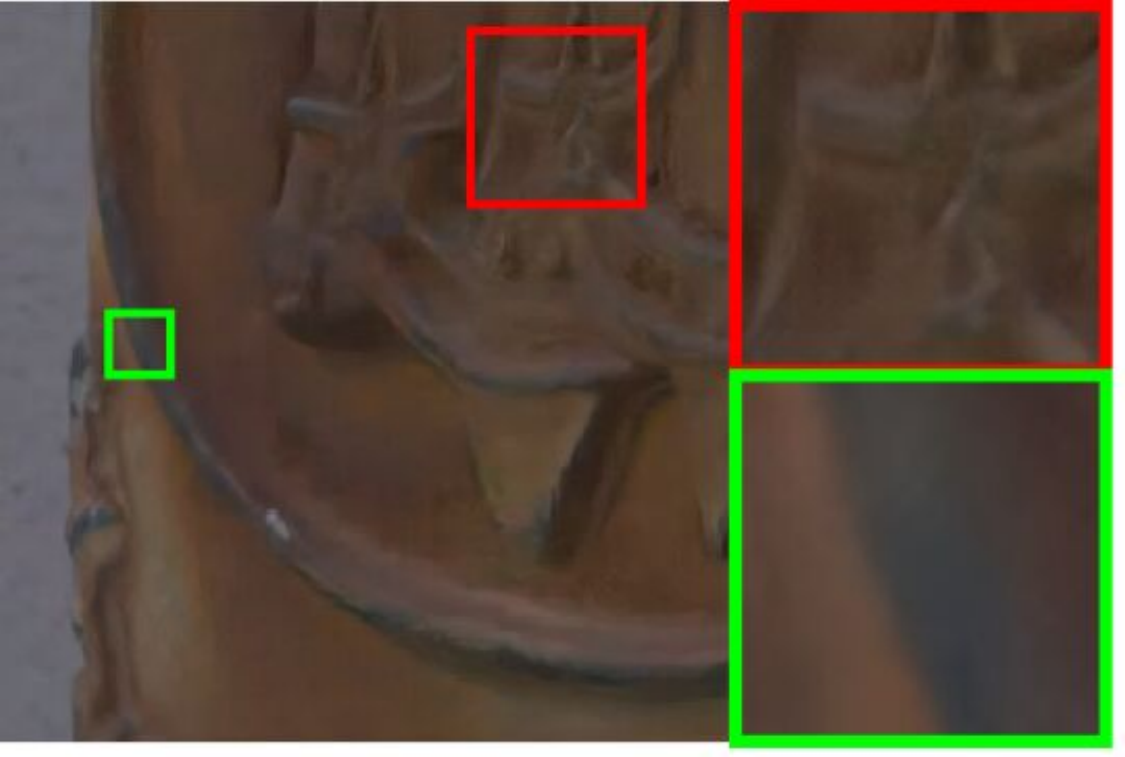}
		\caption{Ours(s=2, k=0) (36.08dB)} 
	\end{subfigure}
	\caption{Denoising results on DND Benchmark. Red box indicates texture details while the green box background or edge.} 
	\label{fig:real2}
\end{figure*}
\paragraph{Mixed AWGN and RVIN.}

\begin{table}[ht]\setlength{\tabcolsep}{2pt}
\centering
\small
\caption{Comparison of PSNR results on mixture of Gaussian noise (AWGN) and Impulse noise (RVIN) removal performance on Set20.}
\begin{tabular}{|c|c|c|c|c|c|c|}
\hline
$(\sigma, r)$ & BM3D & WNNM& PGB & DnCNN-B &Ours-NB & Ours-B\\ \hline 
(10, 0.15)&25.18&25.41&27.17&32.09&\bf32.43&32.37\\
(10, 0.30)&21.80&21.40&22.17&29.97&\bf30.47&30.32\\
(20, 0.15)&25.13&23.57&26.12&29.52&\bf29.82&29.76\\
(20, 0.30)&21.73&21.40&21.89&27.90&\bf28.41&28.16\\
\hline
\end{tabular}
\label{exp:Mix}
\end{table}

Our model follows similar structure of DnCNN and FFDNet~\cite{zhang2018ffdnet}, so its performance on single-type AWGN removal is also similar to them. We thus evaluate our model on eliminating mixed AWGN and RVIN on Set20 as in \cite{xu2016patch}. We also compare our method with other baselines, including BM3D \cite{dabov2006image} and WNNM~\cite{gu2014weighted} which are non-blind Gaussian denoisers anchored with a specific noise level estimated by the approach provided in \cite{liu2013single}. Besides, we include the PGB~\cite{xu2016patch} denoiser that is designed for mixed AWGN and RVIN. The result of the blind version of DnCNN-B, trained by the same strategy as our model, is also presented for reference. The comparison results are shown in Table \ref{exp:Mix}, from which we can see the proposed method achieves the best performance. Compared to DnCNN-B, for complicated mixed noises, our model explicitly disentangles different noises. It benefits the conditional denoiser to differentiate mixed noises from other types. 
\paragraph{Signal-dependent Spatially-variant Noise.}

\begin{table}[t]\setlength{\tabcolsep}{5pt}
\centering
\small
\caption{Comparison of PSNR results on Signal-dependent Noises on CBSD68.}
\begin{tabular}{|c|c|c|c|c|c|}
\hline
$(\sigma_s, \sigma_c)$ & BM3D &  FFDNet &DnCNN-B & CBDNet &Ours-B\\ \hline 
(20, 10)&29.09&28.54&34.38&33.04&\bf34.75\\
(20, 20)&29.08&28.70&\bf31.72&29.77&31.32\\
(40, 10)&23.21&28.67&32.08&30.89&\bf32.12\\
(40, 20)&23.21&28.80&30.32&28.76&\bf30.33\\
\hline
\end{tabular}
\label{exp:SDN}
\end{table}

We conduct experiments to examine the generalization ability of fully convolutional model on signal-dependent noise model~\cite{guo2018toward,foi2008practical,liu2014practical}. Given a clean image $x$, the noises in the noisy observation $y$ contain both signal-dependent components with variance $x\sigma_s^2$ and independent components with variance $\sigma_c^2$. Table \ref{exp:SDN} shows that for non-blind model like BM3D and FFDNet, only scalar noise estimator \cite{liu2013single} is applied, thus they cannot well cope with the spatially-variant cases. In this experiment, DnCNN-B is the original blind model trained on AWGN with $\sigma$ ranged between 0 and 55. It shows that spatially-variant Gaussian noises can still be handled by fully convolutional model trained with spatially-invariant AWGN \cite{zhang2018ffdnet}. Compared to DnCNN-B, the proposed network explicitly estimates the pixel-wise map to make the model more flexible and possible for real noise adaptation.  
\subsection{Evaluation with Real RGB Noise}
\paragraph{Qualitative Comparisons.}
Some qualitative denoising results on DND are shown in Figure~\ref{fig:real2}. The compared results of DND are all directly obtained online from the original submissions of the authors. The methods we include for the comparison cover blind real denoisers (CBDNet, NI~\cite{NI} and NC~\cite{lebrun2015noise}), blind Gaussian denoisers (CDnCNN-B) and non-blind Gaussian denoisers (CBM3D, WNNM~\cite{gu2014weighted}, and FFDNet). From these example denoised results, we can observe that some of them are either noisy (as in DnCNN and WNNM), or spatially-invariantly over-smoothed (as in FFDNet). CBDNet performs better than others but it still suffers from blur edges and uncleaned background. Our proposed method (PD) achieves a better spatially-variant denoising performance by smoothing the background while preserving the textural details in a full blind setting. 

\paragraph{Quantitative Results on DND Benchmark.}
The images in the DND benchmark are captured by digital camera and demosaiced from raw sensor data, so we simply set the stride number $s=2$. We follow the submission guideline of DND dataset to evaluate our algorithm. Recently, many learning-based methods like Path-Restore~\cite{yu2019path},RIDNet~\cite{anwar2019real},WDnCNN~\cite{zhao2019enhancement} and CBDNet, achieved promising performance on DND, but they are all finetuned on real noisy images, or use prior knowledge in the meta-data of DND~\cite{brooks2019unprocessing}. For fair comparison, we select some representative conventional methods(MCWNNM, EPLL, TWSC, CBM3D), and learning-based methods trained only with synthetic noises. The results are shown in Table~\ref{exp:DND}. Models trained on AWGN (DnCNN, TNRD, MLP) perform poorly on real RGB noises mainly due to the large gap between AWGN and real noise. CBDNet improves the results significantly by training the deep networks with artificial realistic noise model. Our AWGN-RVIN-trained model with PD refinement achieves much better results (+0.83dB) than CBDNet trained only with synthetic noises, and also boosts the performance of other AWGN-based methods (+PD). Compared to the base model, the proposed adaptation methods improve the performance on real noises by 5.8 dB. Note that our model is only trained on synthetic noises, and does not utilize any prior data of DND. 


\begin{table}[t]\setlength{\tabcolsep}{1pt}
\centering
\caption{Comparison of PSNR and SSIM on DND Benchmark. PD: Pixel-suffle Down-sampling Strategy. Among all models trained only with synthetic data.}
\begin{tabular}{|l|c|c|}
\hline
Method &PSNR&SSIM \\ \hline 

MCWNNM\cite{xu2017multi}  &37.38 &0.929\\
EPLL\cite{zoran2011learning} & 33.51 &0.824\\
TWSC\cite{xu2018trilateral} &37.93&0.940\\
MLP\cite{burger2012image} & 34.23 &0.833\\
TNRD\cite{chen2017trainable}   &33.65& 0.830\\
CBDNet(Syn)\cite{guo2018toward}  &37.57 &0.936\\ 
\hline
CBM3D\cite{dabov2008image}  & 34.51 &0.850\\
CBM3D(+PD)  & \emph{35.02} &\emph{0.873}\\
\hline
CDnCNN-B\cite{zhang2017beyond} & 32.43 &0.790\\
CDnCNN-B(+PD) & \emph{35.44} &\emph{0.876}\\
\hline
FFDNet\cite{zhang2018ffdnet} & 34.40 &0.847\\
FFDNet(+PD)& \emph{37.56}&\emph{0.931}\\
\hline
Our Base Model(No PD) & 32.60&0.788\\
Ours(Full Pipeline) & \bf38.40&\bf0.945\\
\hline
\end{tabular}
\label{exp:DND}
\end{table}
\subsection{Ablation Study on Real RGB Noise}

\begin{figure}[t]
	\centering
	\begin{subfigure}{0.32\linewidth} 
		\includegraphics[width=\textwidth]{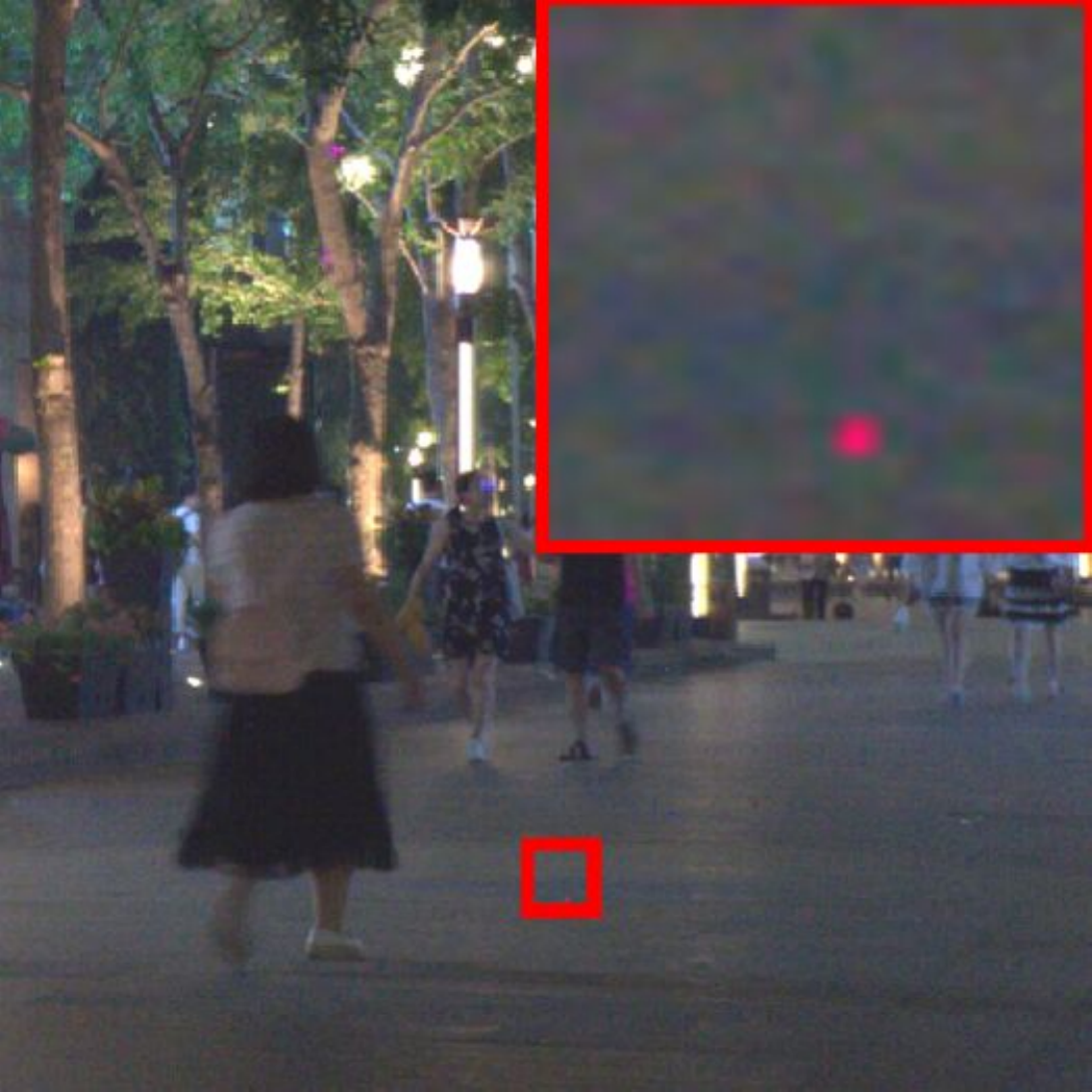}
		\caption{Noisy image} 
	\end{subfigure}
	\begin{subfigure}{0.32\linewidth} 
		\includegraphics[width=\textwidth]{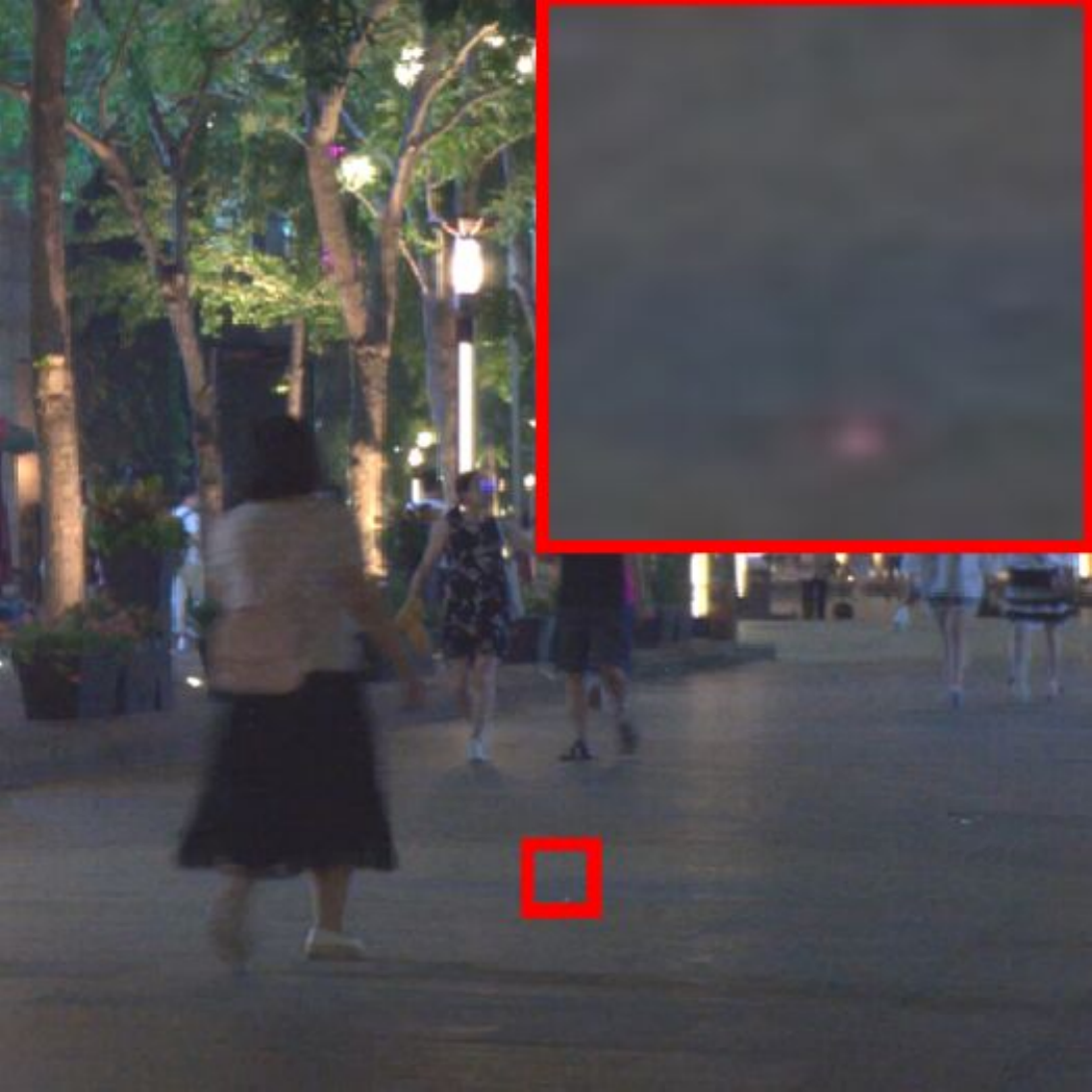}
		\caption{AWGN only} 
	\end{subfigure}
		\begin{subfigure}{0.32\linewidth} 
		\includegraphics[width=\textwidth]{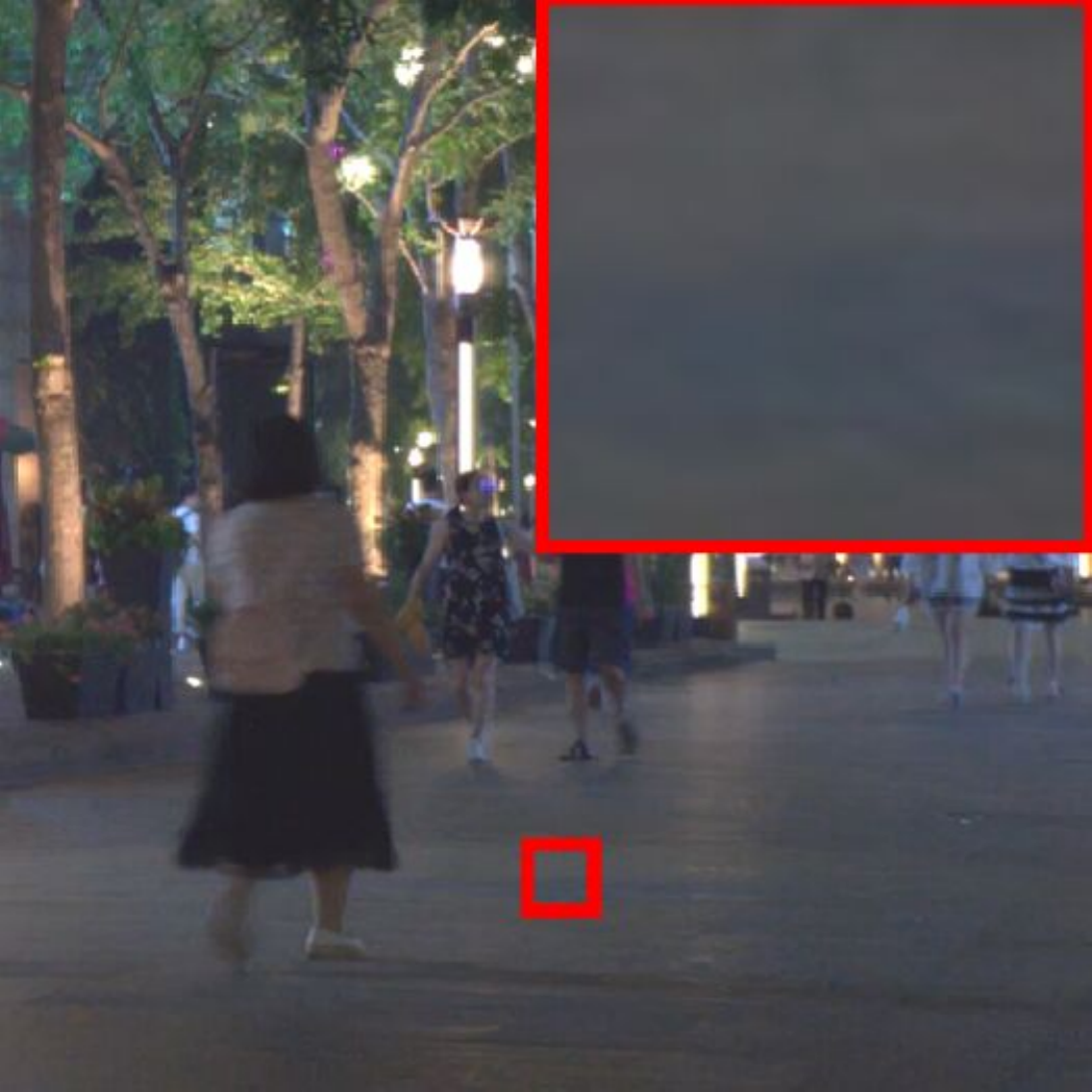}
		\caption{AWGN-RVIN} 
	\end{subfigure}
	\caption{Denoised performance of models trained with AWGN in (b) and mixed AWGN-RVIN in (c). During testing, $k=0$ and $s=2$.  }
	\label{fig:RVIN_ab}
\end{figure}

\paragraph{Adding RVIN.}
Training models with mixed AWGN and RVIN noises will benefit the removal of dead or over-exposure pixels in real images. For comparison, We train another model only with AWGN, and test it on real noisy night photos. An example utilizing the full pipeline is shown in Figure~\ref{fig:RVIN_ab}, in which it demonstrates the superiority of the existence of RVIN in the training data. Even though model trained with AWGN can also achieve promising denoising performance, it is not effective on dead pixels. 
\begin{figure}[t]
	\centering
	\begin{subfigure}{0.24\linewidth} 
		\includegraphics[width=\textwidth]{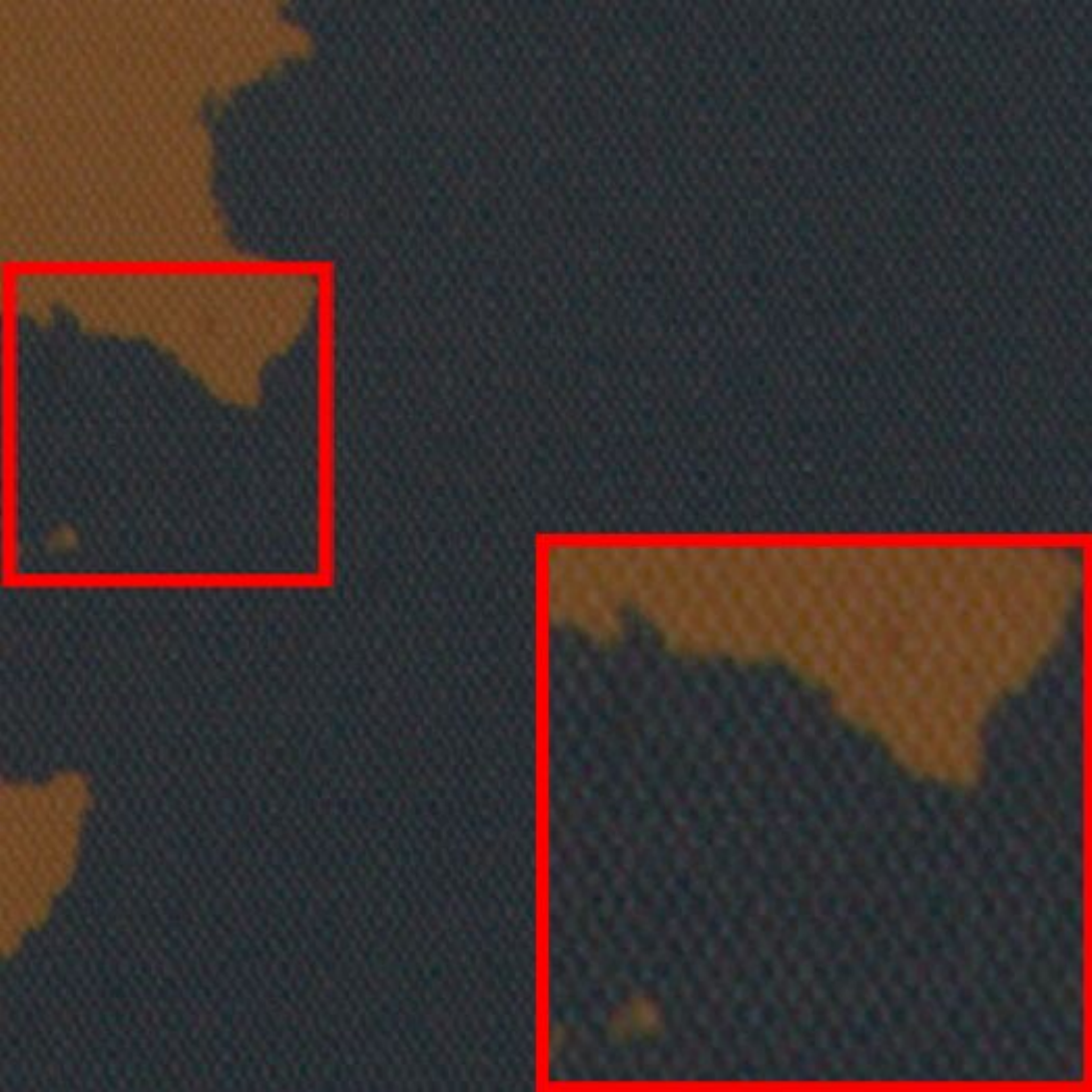}
		\caption{Noisy image} 
	\end{subfigure}
	\begin{subfigure}{0.24\linewidth} 
		\includegraphics[width=\textwidth]{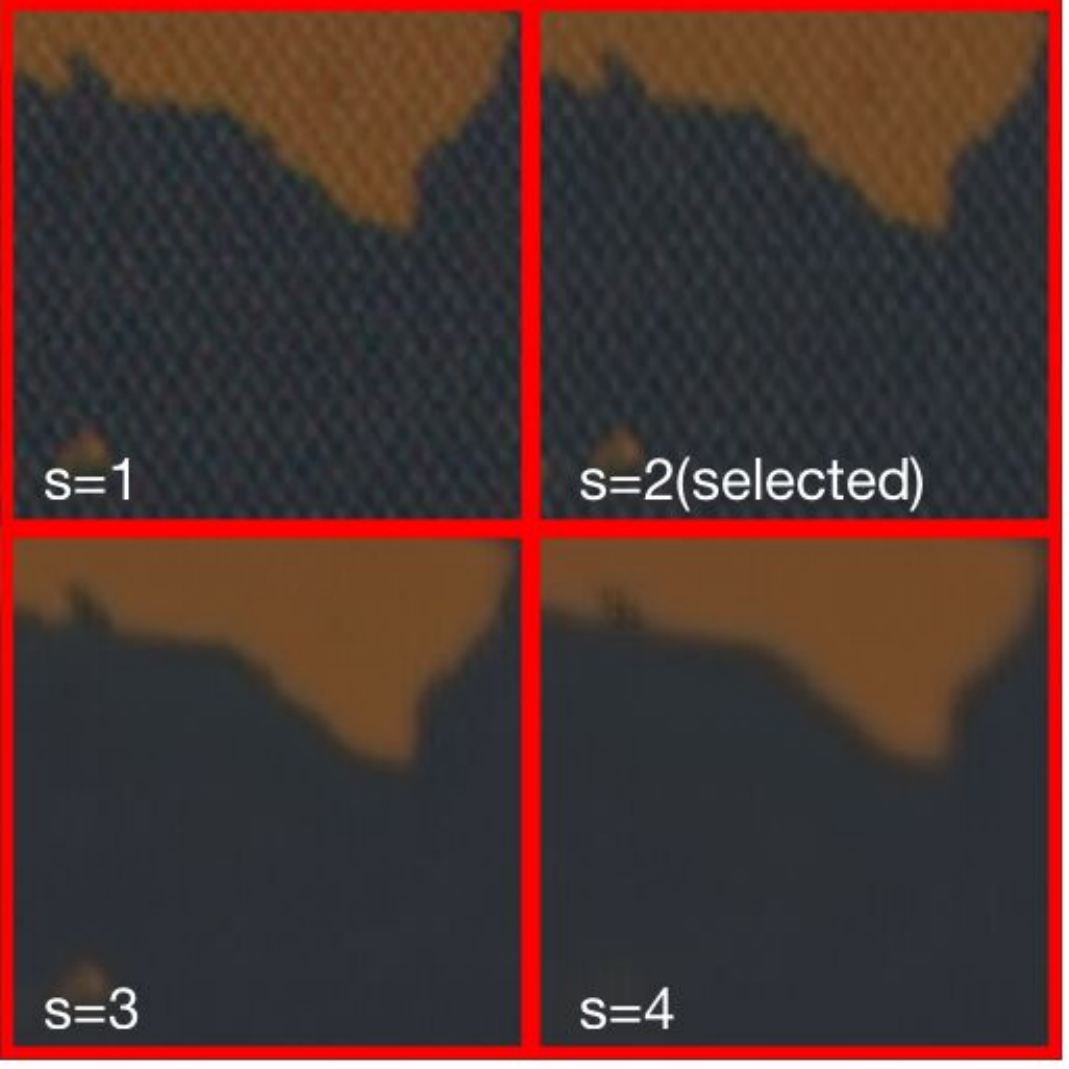}
		\caption{Denoised.} 
	\end{subfigure}
	\begin{subfigure}{0.24\linewidth} 
		\includegraphics[width=\textwidth]{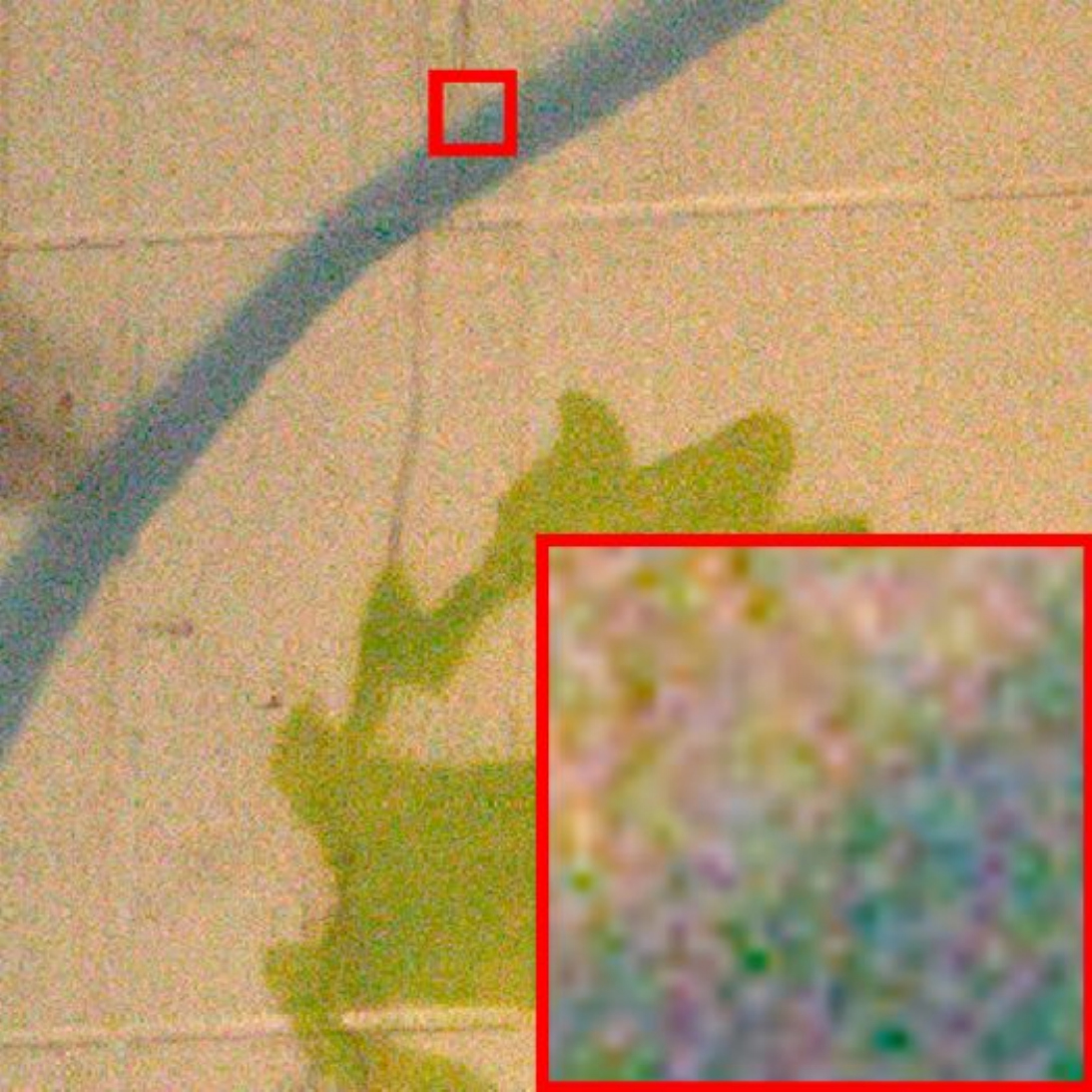}
		\caption{Noisy Image} 
	\end{subfigure}
	\begin{subfigure}{0.24\linewidth} 
		\includegraphics[width=\textwidth]{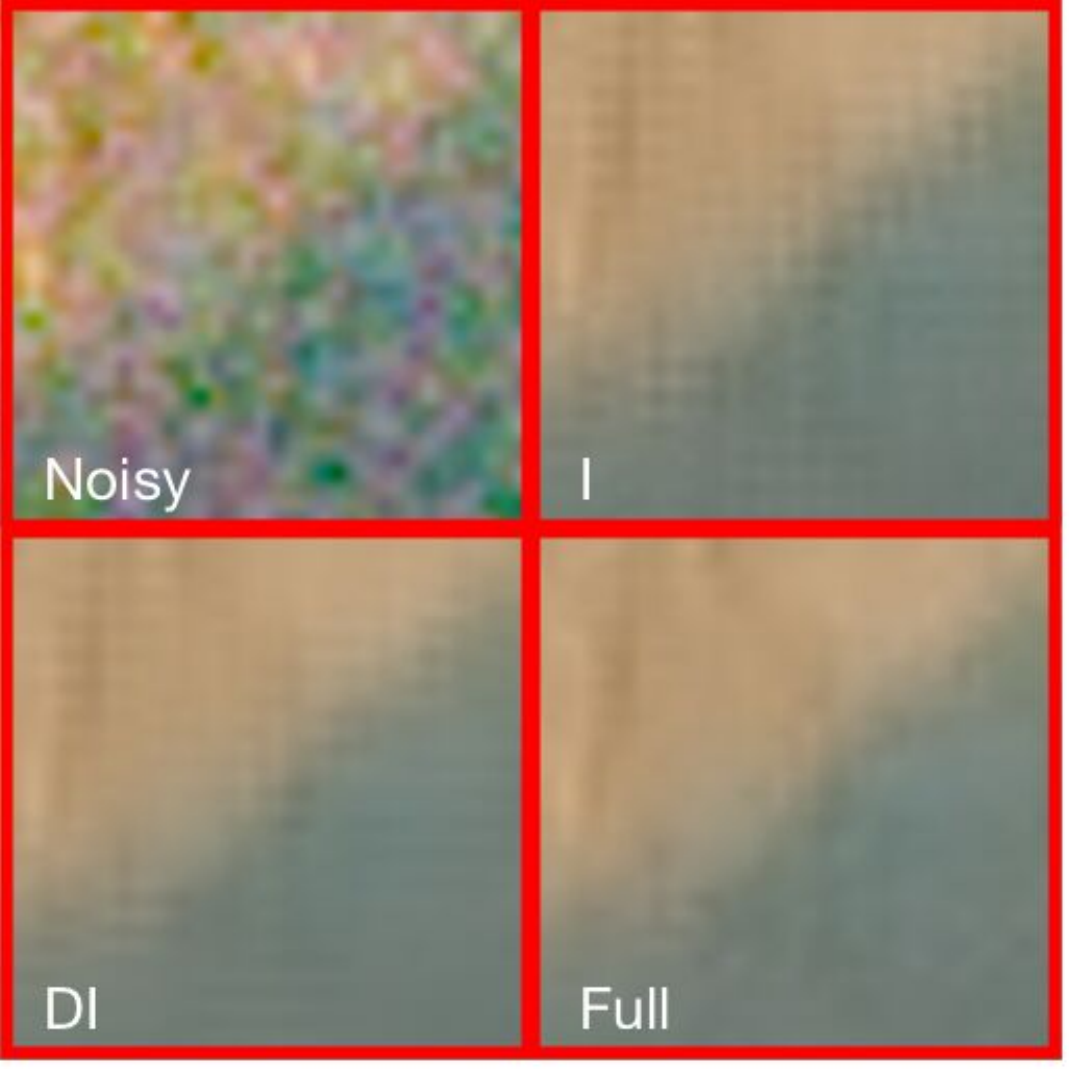}
		\caption{Denoised.} 
	\end{subfigure}
	\caption{(a)(b):Denoised performance of different stride $s$ when $k=0$, and (c)(d): Ablation study on refinement. $s=2$ and $k=0$. }
	\label{fig:process_ab}
\end{figure}
\begin{figure}[t]
	\centering
	\begin{subfigure}{0.15\linewidth} 
		\includegraphics[width=\textwidth]{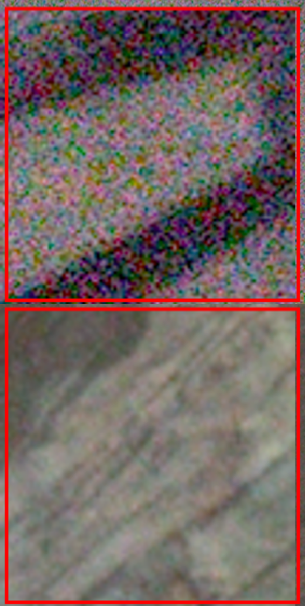}
		\caption{Noisy} 
	\end{subfigure}
	\begin{subfigure}{0.15\linewidth} 
		\includegraphics[width=\textwidth]{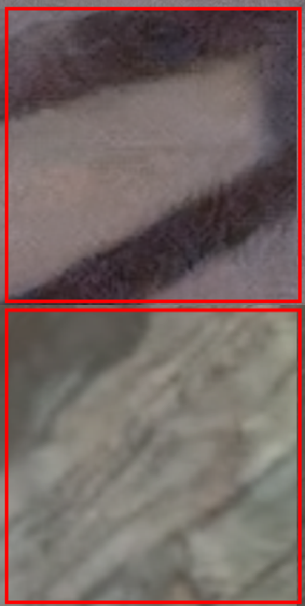}
		\caption{$0$} 
	\end{subfigure}
		\begin{subfigure}{0.15\linewidth} 
		\includegraphics[width=\textwidth]{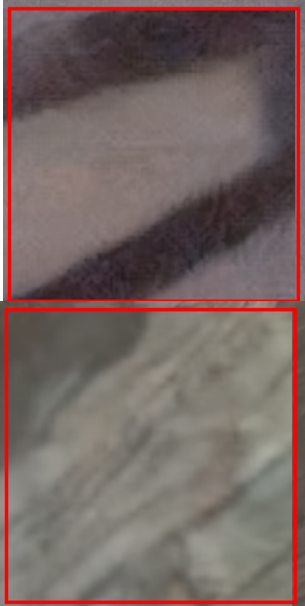}
		\caption{$0.3$} 
	\end{subfigure}
	\begin{subfigure}{0.15\linewidth} 
		\includegraphics[width=\textwidth]{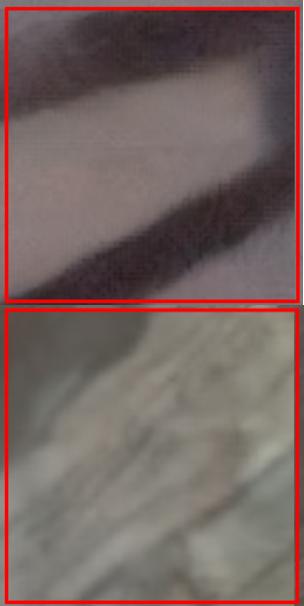}
		\caption{$0.5$} 
	\end{subfigure}
		\begin{subfigure}{0.15\linewidth} 
		\includegraphics[width=\textwidth]{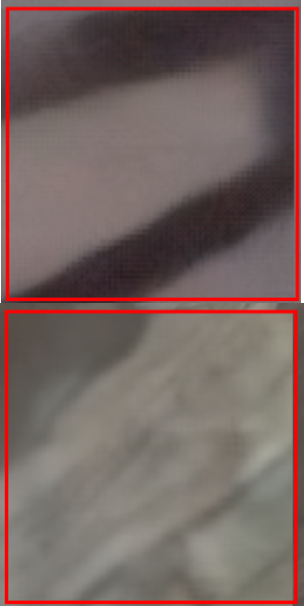}
		\caption{$0.8$} 
	\end{subfigure}
		\begin{subfigure}{0.15\linewidth} 
		\includegraphics[width=\textwidth]{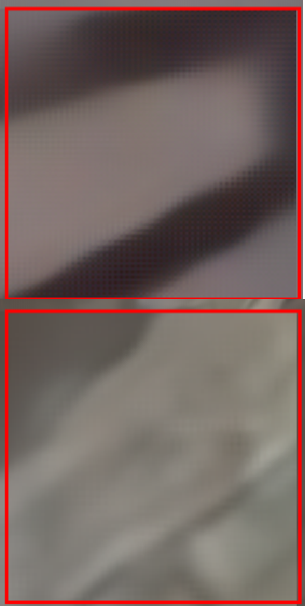}
		\caption{$1$} 
	\end{subfigure}
	\caption{Ablation study on merging factor $k$, and $s=2$.}
	\label{fig:k_ab}
\end{figure}
\paragraph{Stride Selection.}
We apply different stride numbers while refining the denoised results, and compare the visual quality in Figure \ref{fig:process_ab} (a)(b). For arbitrary given sRGB images, the stride number can be computed using our adaptation algorithm with the assistance of noise estimator. In our experiments, the selected stride is the smallest $s$ that $r_s <\tau$. Small stride number will treat large noise patterns as textures to preserve, as shown in Figure \ref{fig:process_ab} (b). While using large stride number tends to break the textural structures and details. Interestingly, as shown in Figure \ref{fig:process_ab} (b), the texture of the fabric is invisible while applying $s>2$. 
\begin{table}[t]\setlength{\tabcolsep}{4pt}
\centering
\small
\caption{Ablation study on refinement steps.}
\begin{tabular}{|c|c|c|c|c|c|}
\hline
Model & (s=1) & (s=3, Full) & (s=2,I) & (s=2,DI) & (s=2,Full) \\ \hline 
PSNR&32.60&37.90&37.00&37.20&\bf38.40\\
SSIM&0.7882&0.9349&0.9339&0.9361&\bf0.9452\\
\hline
\end{tabular}
\label{exp:dndab}
\end{table}
\paragraph{Image Refinement Process.}
The ablation on the refinement steps is shown in Figure \ref{fig:process_ab} (c)(d) and Table~\ref{exp:dndab}, in which we compare the denoised results of I (i.e. directly pixel-shuffling upsampling after step (2)), DI (i.e. denoising I using $\mathcal{G}$), and Full (i.e. the current whole pipeline). It shows that both I and DI will form additional visible artifacts, while the whole pipeline smooths out those artifacts and has the best visual quality.  

\paragraph{Blending Factor $k$.}
Due to the ambiguity nature of fine texture and mid-frequent noises, human perception intervene on the denoising level is inevitable. $k$ is this parameter introduced as a 'linear' adjustment of denoising level for a more flexible and interactive user operation. Using blending factor $k$ is more stable and safe to preserve the spatially-variant details than directly adjusting the estimated noise level like CBDNet. In Figure \ref{fig:k_ab}, as $k$ increases, the denoised results tend to be over-smoothed. This is suitable for images with more background patterns. However, smaller $k$ will preserve more fine details which are applicable for images with more foreground objects. 
In most cases, users can simply set $k$ to 0 to obtain the most detailed textures recovery and visually plausible results.

\section{Conclusions}
In this paper, we revisit the real image blind denoising from a new viewpoint. We assumed the realistic noises are spatially/channel -variant  and correlated, and addressed adaptation from AWGN-RVIN noises to real noises. Specifically, we proposed an image blind and non-blind denoising network trained on AWGN-RVIN noise model. The network consists of an explicit multi-type multi-channel noise estimator and an adaptive conditional denoiser. To generalize the network to real noises, we investigated Pixel-shuffle Down-sampling (PD) refinement strategy. We showed qualitatively that PD behaves better in both spatially-variant denoising and details preservation. Results on DND benchmark and other realistic noisy images demonstrated the newly proposed model with the strategy are efficient in processing spatial/channel variance and correlation of real noises without explicit modeling. 

\bibliography{egbib}
\bibliographystyle{aaai}
\end{document}